\documentclass[letterpaper, 10 pt, conference]{ieeeconf}  

\IEEEoverridecommandlockouts                              

\usepackage{graphicx}

\title{\LARGE \bf
Midas: A Multi-Joint Robotics Simulator with Intersection-Free Frictional Contact
}

\author{Yunuo Chen$^{1,2,*}$, Minchen Li$^{2}$, Wenlong Lu$^{1}$, Chuyuan Fu$^{1}$, Chenfanfu Jiang$^{2}$
\thanks{$^{1}$Everyday Robots, X The Moonshot Factory, Mountain View, CA, USA}%
\thanks{$^{2}$University of California, Los Angeles, Los Angeles, CA, USA}%
\thanks{*Work done as an AI Resident at Everyday Robots.}
}

\begin{document}

\maketitle
\thispagestyle{empty}
\pagestyle{empty}

\begin{abstract}

We introduce Midas, a robotics simulation framework based on the Incremental Potential Contact (IPC) model. Our simulator guarantees intersection-free, stable, and accurate resolution of frictional contact.
We demonstrate the efficacy of our framework with experimental validations on high-precision tasks and through comparisons with Bullet physics.
A reinforcement learning pipeline using Midas is also developed and tested to perform intersection-free peg-in-hole tasks.
\end{abstract}

\section{Introduction}

Frictional contact is ubiquitous in robotics. Contact modeling greatly affects the quality of robotics simulation, and consequently control policies. However, none of the traditional simulators can guarantee accurate resolution of contact. In contact-rich robot manipulation tasks, inaccurate collision models may lead to artifacts such as interpenetration between objects, which can lead to disastrous results when transferred to real robots.

The recently proposed Incremental Potential Contact (IPC) model \cite{li2020incremental}, an implicit time-stepping algorithm based on Finite Element Method (FEM), has been proved to guarantee the resolution of frictional contact for solid elastic objects. IPC has also been extended to simulate codimensional objects \cite{li2020codimensional}, rigid bodies \cite{ferguson2021intersection, lan2022affine} and articulated bodies \cite{chen2022unified}, etc.

Originating from computer graphics, IPC has shown great potential in robotics simulation as well. Kim et al. \cite{kim2022ipc} adopted IPC to simulate parallel-jaw robot grippers and demonstrated that IPC can significantly narrow the sim-to-real gap for grasping tasks.

In this paper, we propose a general robotics simulation framework incorporating the recent advancements of articulated rigid body simulation using IPC formulation.
Our simulator has guaranteed numerical accuracy and is suitable for various robot control tasks, especially the challenging contact-rich ones.

The contributions of this paper include:
\begin{itemize}
    \item Midas: a robotics simulation framework with guaranteed non-interpenetration, strong numerical stability and accurate motor controls.
    \item Experimental validations on high-precision tasks and comparison with the Bullet simulator.
    \item A reinforcement learning pipeline using Midas to perform intersection-free peg-in-hole tasks.
\end{itemize}

\section{Related Work}

\subsection{Contact Simulation and IPC}
Contact simulation for rigid bodies has been extensively studied in robotics and graphics. Early rigid body contact dates back to simple spring-damper models \cite{lankarani1989contact, rosenberg1993perceptual}, resulting in stiff systems or there can be large penetrations.

Later, rigid body contacts are mostly formulated as the Linear Complementarity Problem (LCP) \cite{baraff1994fast, stewart1996implicit}, which has been adopted by various physics engines, including ODE \cite{ode}, PhysX \cite{physx} and Bullet \cite{coumans2015bullet}. Todorov et al. \cite{todorov2010implicit} introduced the implicit complementarity solver and incorporated it into the MuJoCo physics engine \cite{todorov2012mujoco}.

In 2020, Li et al. \cite{li2020incremental} proposed the Incremental Potential Contact model based on optimization time integration. IPC computes the exact distance for each contact pair and uses a smoothly clamped barrier function to rewrite contact constraints as energy terms in the objective function. At each optimization step, Continuous Collision Detection (CCD) is performed to strictly prevent interpenetration.

The original IPC formulation treats all objects in the simulator as deformable FEM meshes. Hence, directly using IPC to simulate rigid bodies is inefficient. Several works have been conducted to accelerate the simulation of stiff objects in the IPC framework.
Ferguson et al. \cite{ferguson2021intersection}
introduces the standard rigid body degrees of freedoms (DOF) with rotations and translations into the IPC framework. However, such DOF reduction will result in curved optimization trajectories, which would need the time-consuming curved CCD in contact handling. To avoid this issue, Lan et al. \cite{lan2022affine} proposed Affine Body Dynamics (ABD), which simulates stiff objects as affine bodies with translation and affine deformation DOFs.
In ABD formulation, a rigid body will have 12 DOFs. This is slightly larger than a standard rigid body model, but such a compromise keeps the optimization trajectories linear, which allows the efficient linear CCD to be applied.

More recently, general articulation constraints are introduced to IPC framework \cite{chen2022unified}. The constraints include linear equality constraints (e.g., connect two points together), nonlinear equality constraints (e.g., fix the distance between two simulated objects) and inequality constraints (e.g., bound the rotation range of a hinge joint). The linear equality constraints are exactly enforced using change-of-variables, while the other nonlinear equality and inequality constraints are resolved by adding extra penalty terms to the objective. These articulation constraints provide us with essential ingredients to replicate a robot with multiple joints in IPC simulation.

\subsection{Robot Manipulation for Contact-rich Tasks}
With advanced control techniques, modern robots are installed to conduct complicated tasks with high-precision contacts. Reinforcement Learning (RL) has shown great potential in robot control. However, no single RL algorithm suffices for all robot manipulation tasks. Controlling robots in contact-rich tasks usually requires extra work to meet the demand for high precision.

Various modifications have been proposed to improve the efficiency of RL in specific tasks. 
Beltran et al. \cite{beltran2020variable} proposed an off-policy model-free method to solve peg-in-hole tasks
with hole-position uncertainty. 
Xu et al. \cite{xu2020cocoi} proposed to use online context inference to embed dynamics properties in RL for non-planar pushing tasks. 
Shi et al. \cite{shi2021proactive} incorporated  operational space visual and haptic information into RL to improve policy learning in assembly tasks such as RAM insertion.
Luo et al. \cite{luo2021learning} proposed robotless assembly environments which take force/torque in task space as observation.
Ichiwara et al. \cite{ichiwara2022contact} chose to use vision and tactility to improve performance for bag unzipping tasks.
Narang et al. \cite{narang2022factory} explored contact reduction in assembly tasks and achieved real-time simulation of a wide range of contact-rich scenes.
Levin et al. \cite{levine2015learning} proposed a policy search method to learn a wide range of manipulation behaviors.
Spector et al. \cite{spector2020deep} incorporated pruning methods to facilitate convergence for peg-in-hole tasks.

\section{Simulator Design}
In this section, we discuss the design details of our simulator. 

\subsection{Overview}

We follow standard IPC to formulate each simulation step as an optimization problem. At step $n$, the optimization variable is the DOFs $x^n$ of the affine bodies (12 DOFs per body). The objective energy is given as
\begin{equation}\label{eqn:optimize}
    \min_{x^{n+1}}I(x^{n+1}) + h^2\left( \Psi (x^{n+1}) + B(x^{n+1}) + P(x^{n+1}) \right)
\end{equation}
Here $h$ is the current timestep size, $I$, $\Psi$, $B$ are the inertia, elasticity and frictional contact energy respectively. The extra $P$ function summarizes nonlinear equality and inequality articulation constraints. The energy function (\ref{eqn:optimize}) is optimized, where the local minimum gives the configuration of the next step.

The core part of simulating a robot in this framework is to translate robot geometry and connectivity as affine body DOFs and constraints. In our simulator, we support loading a robot described by Unified Robot Description Format (URDF) file, where a robot is defined as a series of links (rigid objects) connected by joints. When creating a robot, we initialize an affine body for each link according to its Cartesian world position to track its motion throughout the simulation. The inertia energy $I$ and elasticity energy $\Psi$ are now computed using the affine DOFs $x^{n}$.
In addition, we feed the exact surface mesh of link geometry to the simulator for computing contact energy $B$. 

Now we have converted robot links to rigid bodies in our simulator, the remaining part is to translate robot joints and motors to the articulation constraints derived in \cite{chen2022unified}. These constraints, both linear and nonlinear, can be consistently resolved in the optimization by either performing a coordinates transformation on the DOFs $x^n$, or adding an extra term into the constraint energy $P$.

\subsection{Joint Constraints}
We support all types of joints described in a URDF file. For most types of robot joints, we can directly find equivalent 
 articulation constraints in \cite{chen2022unified}.   Here we discuss the three most commonly used joints:

A \textit{sliding (prismatic)} joint (such as parallel-jaw gripper) is constructed by adding a relative sliding constraint in \cite{chen2022unified}. Such a constraint will effectively enforce the child link to stay on a straight line in the local coordinates of the parent link.

A \textit{hinge (revolute/continuous)} joint can be imposed by adding two point connections between parent and child link on the hinge axis. Under such restrictions, the child link can only rotate around the axis.

A \textit{fixed} joint can be formed by adding three non-collinear point connections. This can be viewed as adding two orthogonal hinge joints which will eliminate all relative transformation between the two links.

\begin{figure}[thpb]
    \centering
    \includegraphics[scale=0.2]{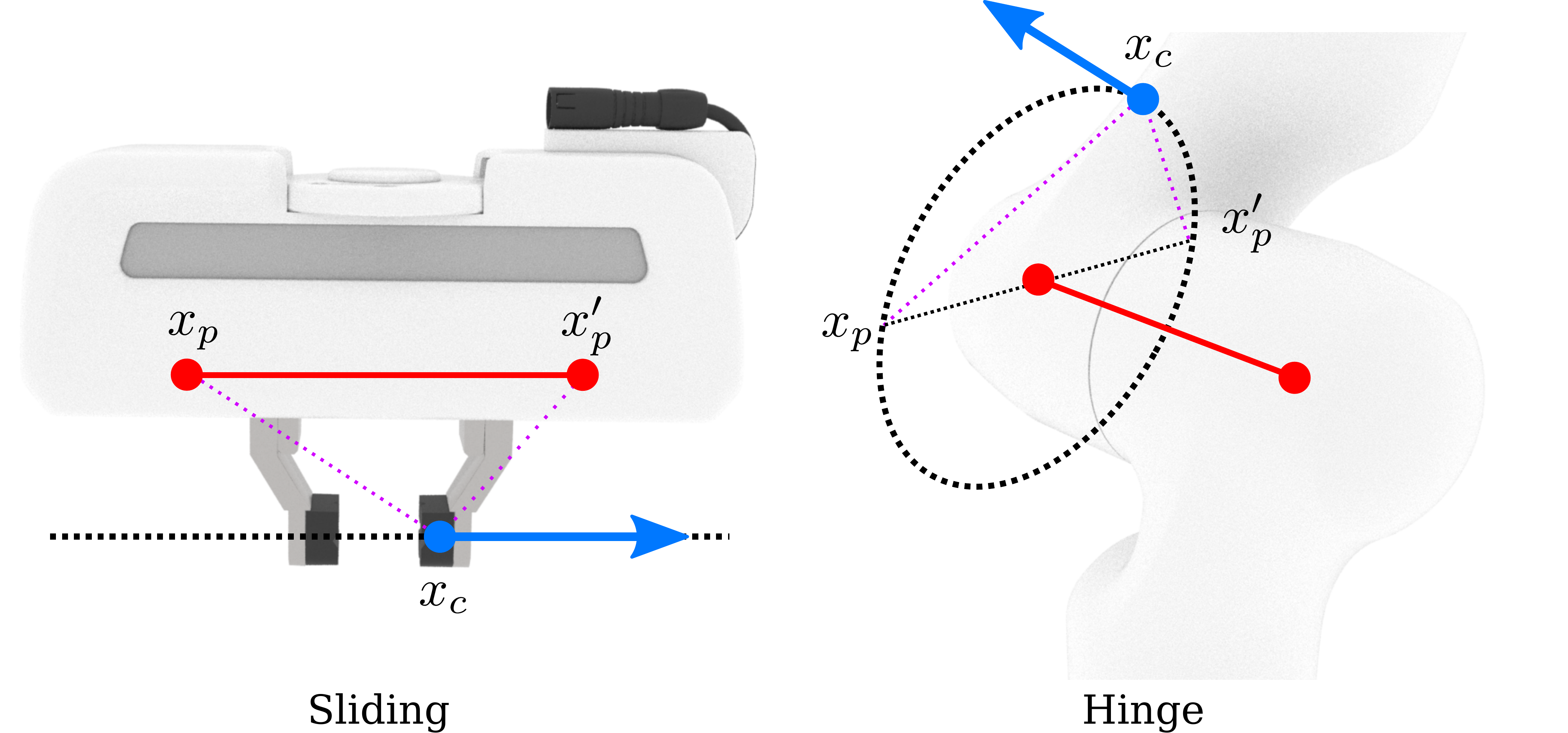}
    \caption{Illustrations of sliding and hinge joints. The red solid line represents the joint axis, while the black dotted line gives the trajectory of the child link. The blue arrow points to the current direction of the joint motor. Additionally, the purple dashed lines indicate distance-based constraints.}
    \label{fig:joint}
\end{figure}

In sliding and hinge joints, we can enforce joint limits to restrict the child link to stay within a given range, instead of moving freely. Such limits can be formulated as bounded distance in \cite{chen2022unified}. For limited joints, we select a point $x_c$ on the child link and a point $x_p$ on the parent link (See Fig. \ref{fig:joint}). We can impose upper (lower) limits on $x_c$ by adding upper (lower) bounds on the distance between $x_p$ and $x_c$. These inequality constraints are then passed to the simulator and resolved using the same barrier technique used in contact.

\subsection{Motor Controls}
Controlling the movement of a robot is essentially decoupled to controlling each joint. We support both position and torque controls for the sliding and hinge joints. Here we briefly introduce how these motors can be written as constraints. 

A position (velocity) motor control moves the child link to a target position. Similar to imposing joint limits, we again use distance-based constraints to enforce this. Specifically, we change the distance inequality constraints between $x_c$ and $x_p$ to equality constraints. Now $x_c$ is fixed at a given position instead of floating around. Note that one single distance equality may lead to multiple solutions on the trajectory of $x_c$. To resolve this issue, one can add an additional point $x_p'$ on the parent link and use two distance equality constraints to realize the motor.

A torque (force) control is an external point force exerted on the child link. In the optimization framework, this can be simply applied by adding external forces (similar to gravity) to the respective affine body DOFs. The force points to the tangential direction of the child link trajectory. For the sliding joint it is the axis direction, while for the hinge joint it is the tangent direction of the current location on the circular trajectory.

\subsection{Simulation}
After loading the robot geometries as affine bodies, treating connectivities and motor controls as constraints, we are able to proceed with our simulation using the optimization time integrator. At each step, we compute the incremental potential (\ref{eqn:optimize}) and locally optimize it using Newton's method with line search. When the simulation is done, we translate the affine body DOFs to Cartesian space and update the robot configurations accordingly.

\section{Simulator Validations}

To demonstrate the efficacy of our simulator, we select a series of robot tasks in which accurate contact is needed to faithfully reproduce reality. All of the experiments are conducted with a Franka Panda parallel-jaw gripper.

\subsection{Assembling Beams}
\begin{figure}[thpb]
    \centering
    \includegraphics[scale=0.15]{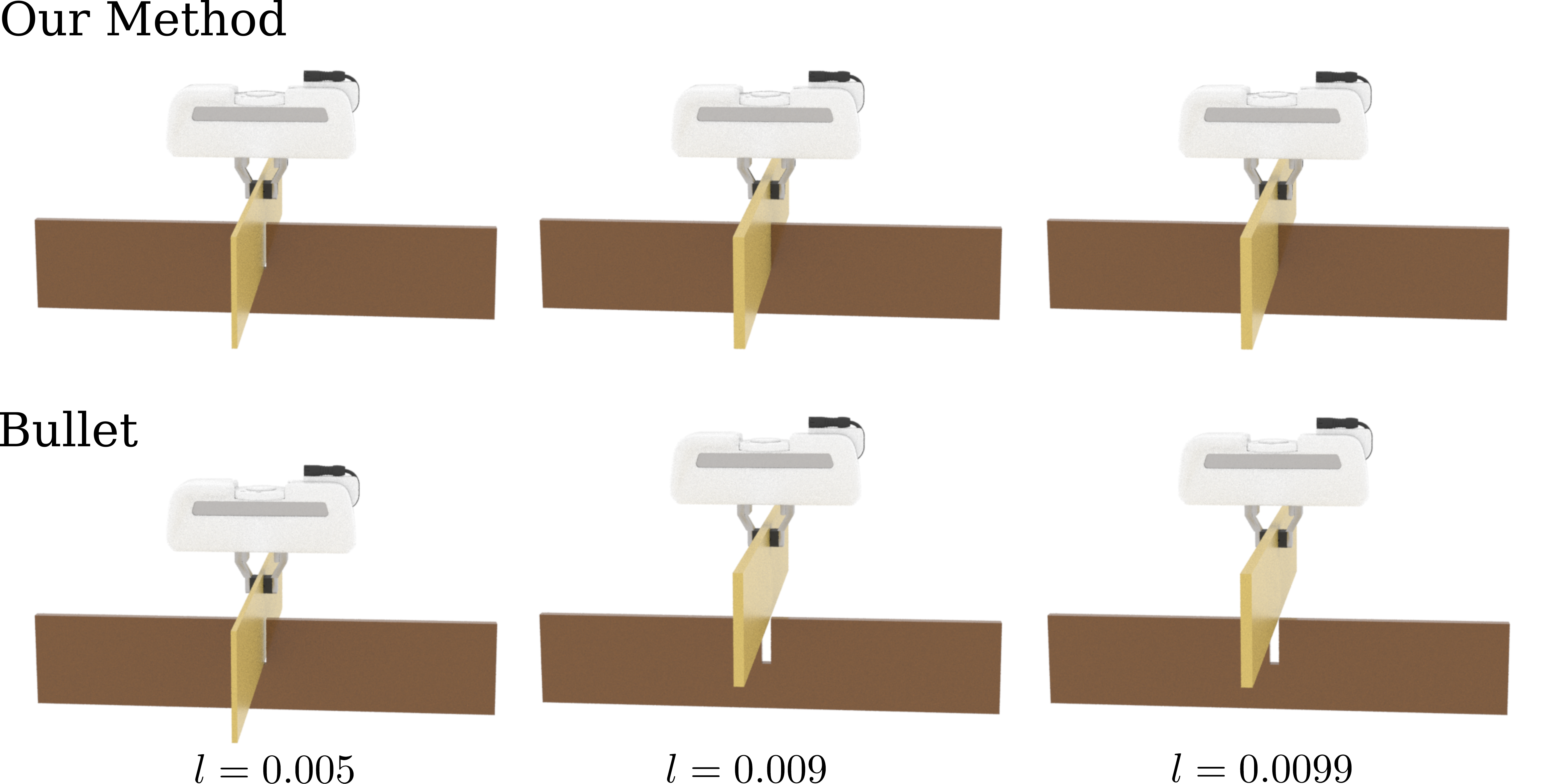}
    \caption{We simulate a robotic assembly task by inserting the top beam into the bottom beam's notch. We choose beams with different widths $l$. Our simulator can successfully finish the task when the margin is less than $0.0001 m$, while Bullet fails when the margin becomes smaller.}
    \label{fig:beam}
\end{figure}

We start by performing a robotic assembly task with two beams. We use a gripper to control the top beam and try to insert it into a notch on the bottom beam. The bottom beam is $0.5 m\times 0.1 m \times 0.01 m$, with a $0.01 m$-width notch in the middle. The top beam is a complete $0.5 m\times 0.1 m \times l m$ cuboid, where $l$ is smaller than $0.01$, hence it can be inserted into the bottom one's notch. In our experiment, we test with three top beams with width $l = 0.005, 0.009, 0.0099$ respectively. Due to the high accuracy of our contact model, we are able to simulate all three insertions without interpenetrations or artificial locking. In fact, by setting the contact clamping threshold $\hat{d}$ smaller, our simulator can simulate arbitrarily small contact distance (e.g. $10^{-7} m$), far beyond the precision needed in most everyday robots.

To compare our method with other simulator, we conduct the same experiment in PyBullet \cite{coumans2021}. Due to the default and unadjustable collision margin of $0.001m$, PyBullet can only insert the first beam successfully. The gripper fails to insert the second and the third ones as the margins between beams are too small.

\subsection{Pick and Place}
\begin{figure}[thpb]
    \centering
    \includegraphics[scale=0.2]{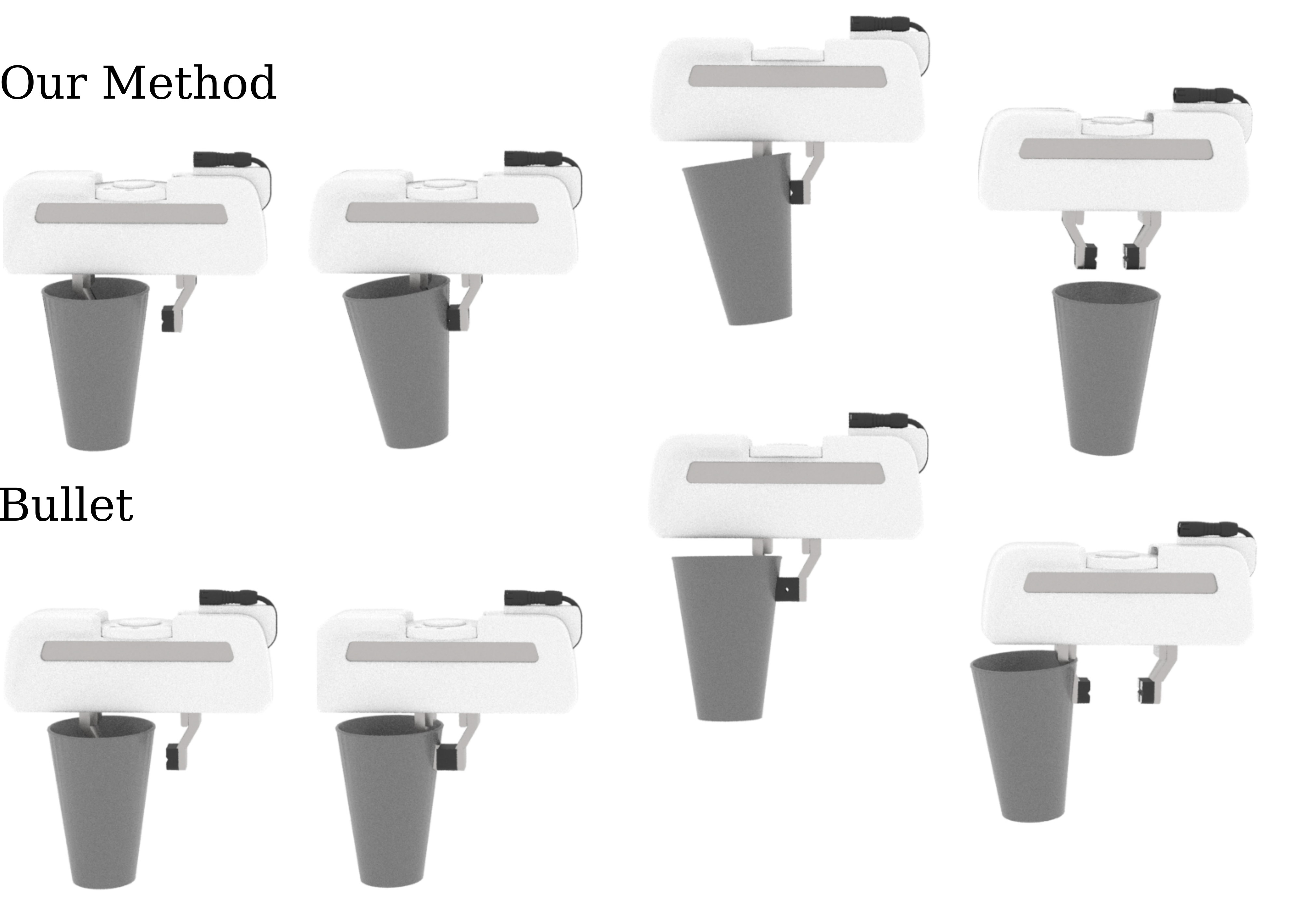}
    \caption{We control the gripper to pick up a cup and place it at a nearby location. Our simulator can successfully perform the task while Bullet fails due to interpenetrations between the gripper and the cup.}
    \label{fig:cup}
\end{figure}

Picking up objects with thin-shell structures such as plates and cups is usually not an easy task in robotics simulation. This is because interpenetrations may occur between objects and grippers.
Here, we conduct our experiment on a thin-shell cup. The thickness of the cup is around $0.002 m$. We use the gripper with a torque motor to pick up the cup and place it at a nearby location. 

Although the mesh of the cup is relatively thin, our gripper is able to move the cup steadily with accurately resolved frictional contact. Even if we apply a much larger force (e.g., $10\times$ the current force) on the gripper, our simulator is still able to resolve the collision between the gripper and the cup without intersection.
However, when simulating this task with the same motor control in Bullet, the left jaw of the gripper is fully penetrated through the cup after grasping, causing the cup to stuck on the gripper. Therefore, the gripper is unable to drop the cup when the parallel jaw opens. This experiment demonstrates how interpenetrations can lead to undesirable results in robotics simulation.

\subsection{Grasping}
We design a simple grasping task to demonstrate our ability to accurately control the forces applied to joints. We use our gripper with torque-based motor control to grasp a cube on the table. Then we move the gripper upwards at a speed of $0.01m/s$. The gripper will fail to grasp the cube if the joint force is not big enough to balance the gravitational force of the cube.

Here we record the smallest torque control necessary to pick up the cube when we vary the cube density. We observe that when other physical parameters are fixed, the torque control is proportional to the mass of the cube. This aligns with our expectation as the torque force is proportional to the static frictional force needed to balance the gravity.

\begin{figure}[thpb]
    \centering
    \includegraphics[scale=0.18]{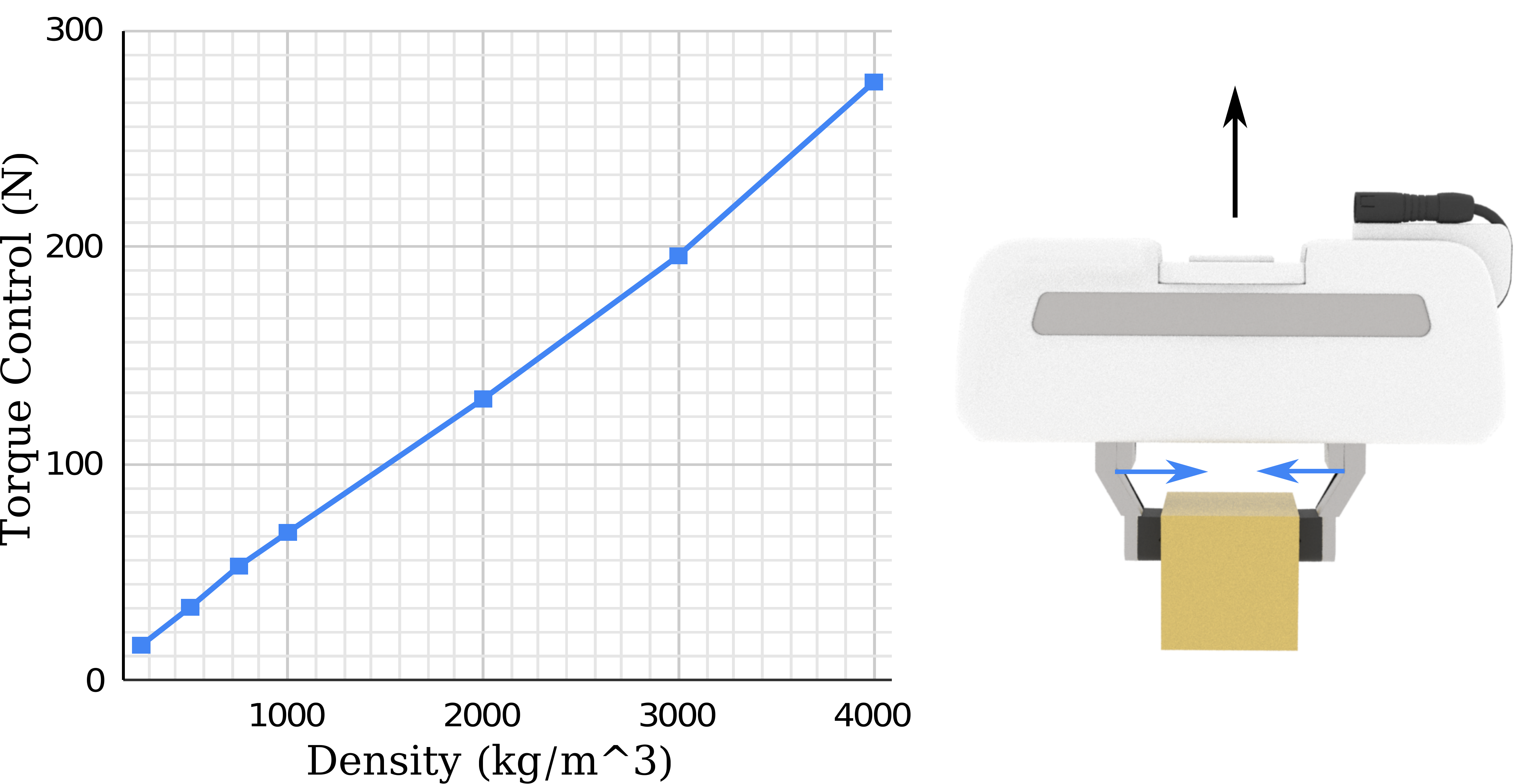}
    \caption{We use torque control to grasp a cube on the table and record the smallest force needed to keep balance. The smallest torque force needed to grasp the cube is proportional to the mass of the cube.}
    \label{fig:grasp}
\end{figure}

Handling contact with IPC also gives us accurate control of friction. When we fix the mass of the cube and vary the friction coefficient $\mu$, we observe a similar relation between friction and torque control, i.e., when $\mu$ is decreased by half, we need to double the joint force in order to keep balance.

\subsection{Inserting USB}
\begin{figure}[thpb]
    \centering
    \includegraphics[scale=0.24]{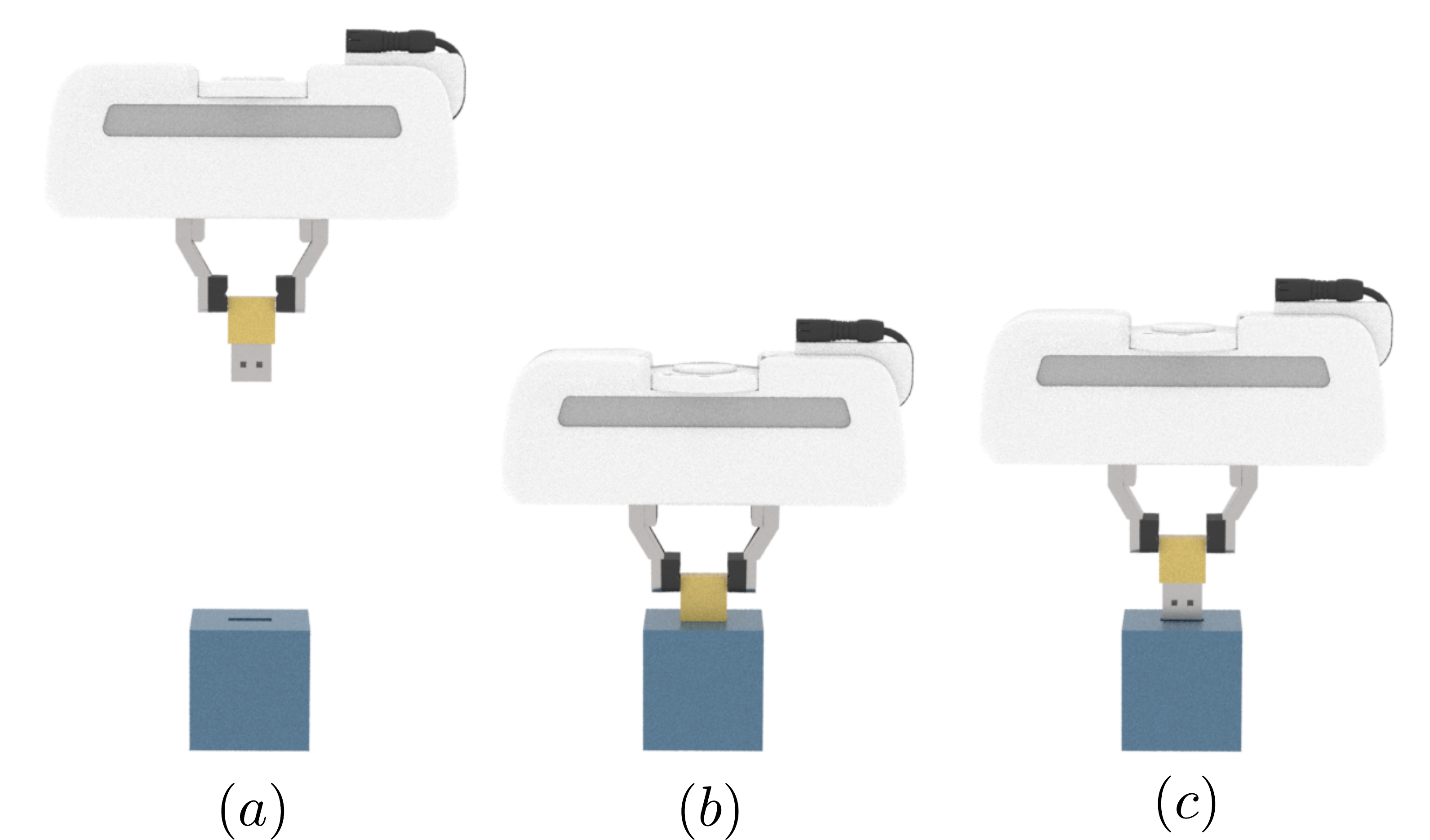}
    \caption{We simulate the insertion of a USB stick (a). Our method can resolve the internal contact (b) while Bullet fails to proceed (c).}
    \label{fig:usb}
\end{figure}

Simulating objects with concave geometries is usually more challenging than dealing with simple convex objects. Hence robotics simulators may find difficulties handling objects with intricate structures such as cables and cords. Here we model a USB-A stick with a corresponding port. Both the USB cord and port have concave geometries and fine internal structures so they can fit tightly when inserted. These fine structures can pose significant challenges to contact handling for the simulators. 

We simulate the insertion process operated by a gripper using our simulator and Bullet (See Fig. \ref{fig:usb}). Our method can accurately model the internal concave geometries and successfully simulate the completion of this task. However, even with a high-resolution convex decomposition, Bullet is not able to insert the USB stick into the port.

\section{Control Policy Learning}

With our guaranteed resolution of contact, we are able to perform contact-rich robot control learning tasks, in which existing simulators may lead to interpenetrations and fails to learn a useful policy when applied to real-world scenarios. In this section, we combine our simulator with reinforcement learning tools to train interpenetration-free control policies for peg-in-hole tasks.

\subsection{Problem Statement}
We use a parallel-jaw gripper to insert a cylinder object (radius $r_1 = 0.035m$, height $h = 0.2m$) into a narrow container. The container has an internal funnel structure with an inner radius ($r_2 = 0.040 m$) slightly larger than the object. We use the difference between radii $\Delta r = r_2 - r_1$ to measure the difficulty of the task.

At the beginning, the gripper is placed at a random interpenetration-free initial position above the funnel. Throughout the simulation, we apply velocity motor control on the gripper to move the object downwards into the funnel. The task is considered complete when the object reaches a target position (depth).

\begin{figure}[thpb]
    \centering
    \includegraphics[scale=0.23]{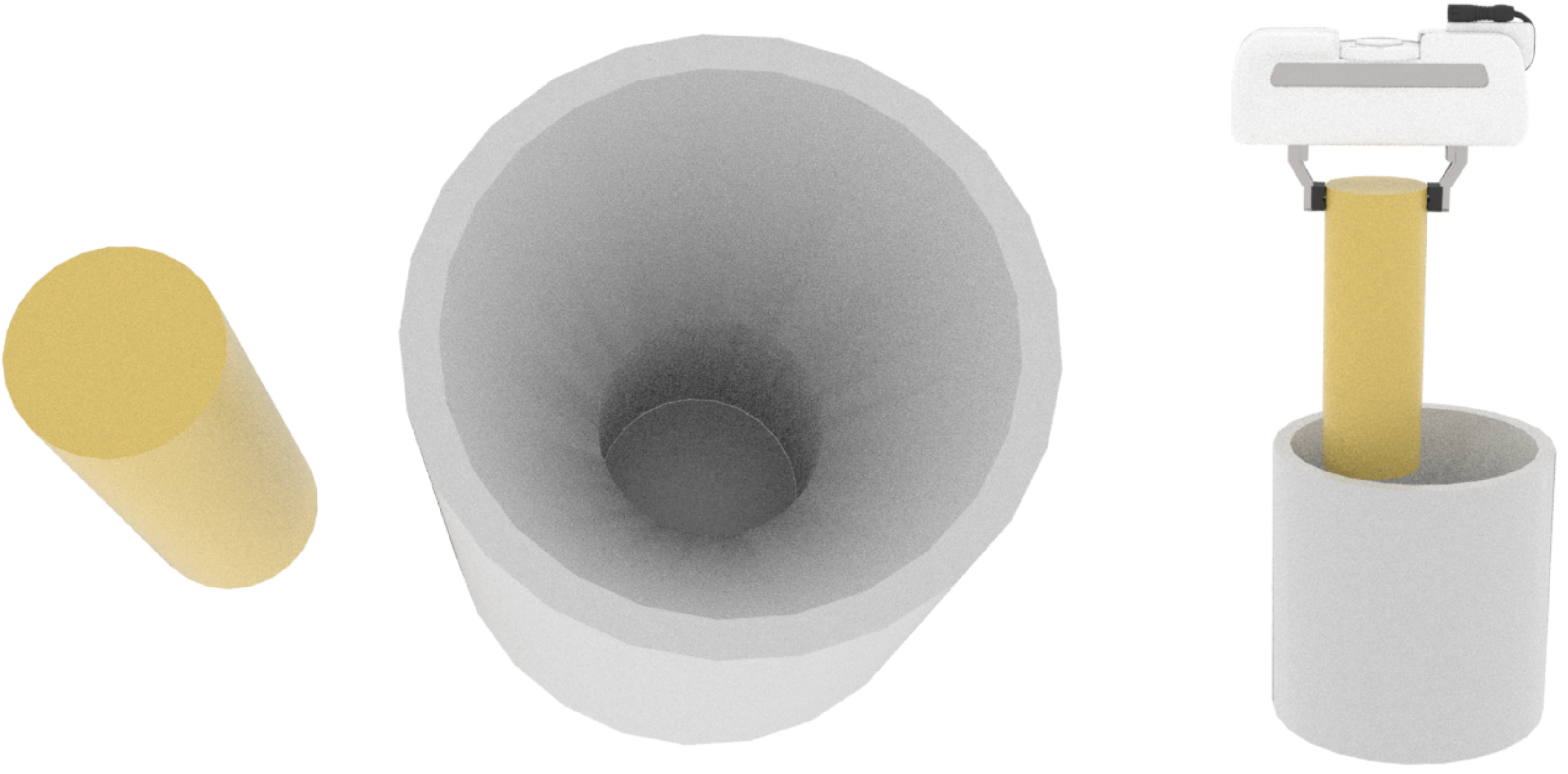}
    \caption{We perform the peg-in-hole task with a cylinder object (left) and a narrow container (middle). Initially, the gripper places the object at a random position above the container (right).}
    \label{fig:peginitial}
\end{figure}

\subsection{Environment}
We construct a continuous control reinforcement learning environment for the proposed peg-in-hole task. Our agent learns the control policy on the gripper joints to insert the object into the container. Here we briefly describe the environment design for this task:
\subsubsection{Action Space}
The action space of the agent is a three-dimension velocity-based motor $(v_1, v_2, v_3)$ on the gripper, which controls the movement of the object.
\subsubsection{Observation Space}
The observation space is the three-dimension Cartesian world position $(x_1, x_2, x_3)$ of the object.
\subsubsection{Reward}
The reward $R$ given to the agent is consist of two parts, $R=R_1+R_2$. $R_1$ is a simple linear reward based on the current velocity of the object in the $z$ direction (agent receives a positive reward if the object is moving downwards inside the container, or a negative reward otherwise). $R_2$ is an additional large reward added right after the agent reaches the target position inside the container. We set $R_2 = 100 + 100 \cdot (1 - s / S)$, where $s$ is the current step and $S$ is the maximum step allowed. Thus the agent will receive a higher reward for finishing the task in fewer steps.

\subsubsection{Termination}
The simulation runs until the object is fully inserted into the container (target depth is reached), or the maximum step $S$ is reached.

\subsection{Method}

We combine our simulator with a continuous control RL algorithm implemented with OpenAI Gym \cite{brockman2016openai} platform. The policy is expressed as a 3-layer fully-connected neural network. We apply a standard Trust Region Policy Optimization (TRPO) method \cite{schulman2015trust} for policy updates. The agent is trained without any prior knowledge.

For comparison, we also perform the training task under the same settings using PyBullet.

\subsection{Policy Training and Evaluation}
\begin{figure}[thpb]
    \centering
    \includegraphics[scale=0.45]{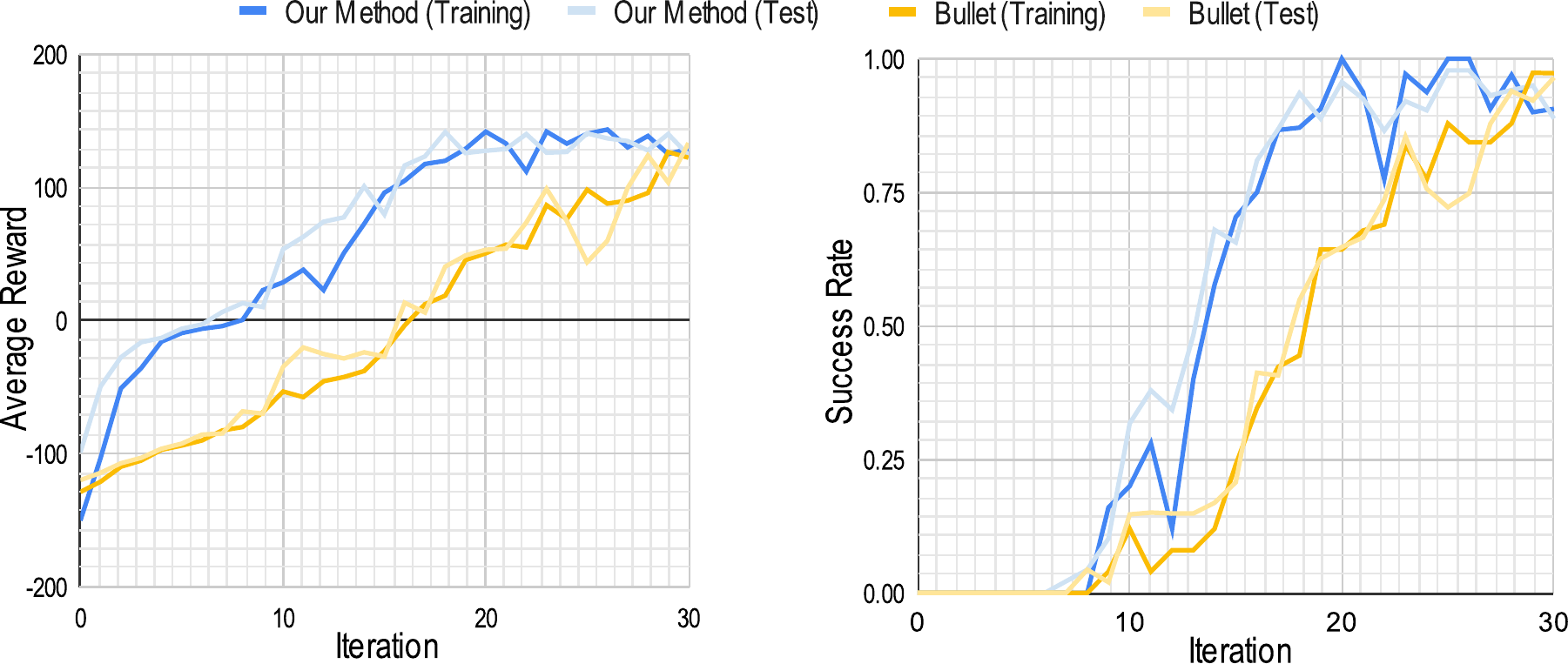}
    \caption{We record the average reward and success rate after each training iteration. Our method achieves faster convergence than Bullet in both reward and success rate.}
    \label{fig:rewardplot}
\end{figure}

Throughout the training process, we regularly test the current trained policy with random test cases. We record the average reward and success rate after each iteration and plot them in Fig. \ref{fig:rewardplot}.

As shown in Fig. \ref{fig:rewardplot}, our simulator outperforms Bullet in terms of reward gain and success rate in the same number of training iterations. After around 20 iterations, the agent trained with our simulator is able to perform the peg-in-hole task from a random initial position, with a success rate of over 90\%, while Bullet achieves a lower success rate (of around 75\%) after the same number of iterations. 

\begin{figure}[thpb]
    \centering
    \includegraphics[scale=0.2]{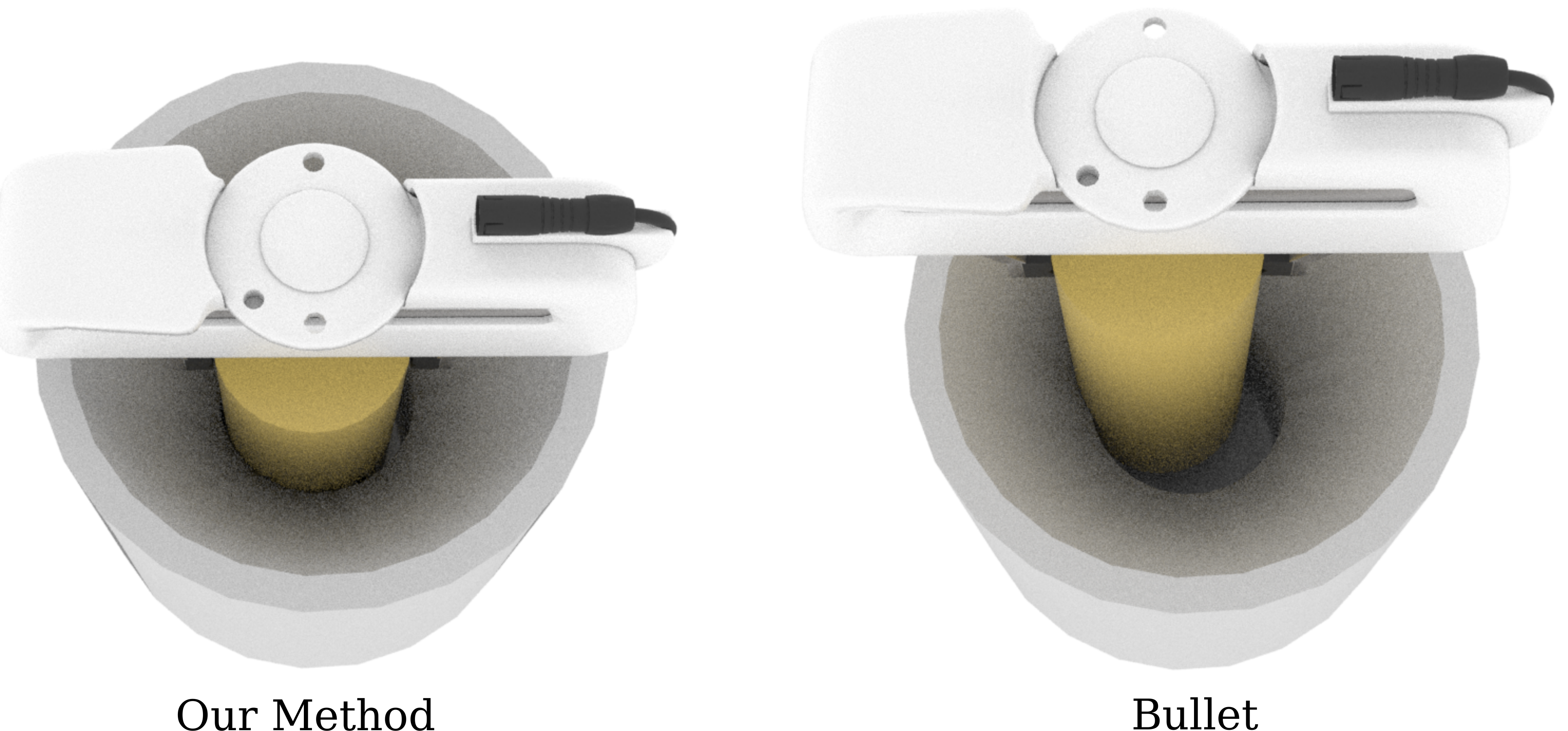}
    \caption{The control policy trained using our simulator is interpenetration-free, while the policy trained using Bullet leads to intersections in test cases.}
    \label{fig:pegfinal}
\end{figure}

Convergence speed is not the only concern for training a peg-in-hole control policy. To examine the quality of the policies, we obtain the agent after 30 training iterations, run tests on random interpenetration-free initial configurations and visualize the results after the simulation is complete (See Fig. \ref{fig:pegfinal}). We observe that, in many of the successful cases, intersections between the object and the container can be easily spotted in the tasks performed by Bullet, while all test cases in our simulator remain interpenetration-free.

To further quantify the quality of the policies, we perform 200 tests on the trained agent. For each test case, we record the final position of the object if it is inserted successfully.

\begin{figure}[thpb]
    \centering
    \includegraphics[scale=0.7]{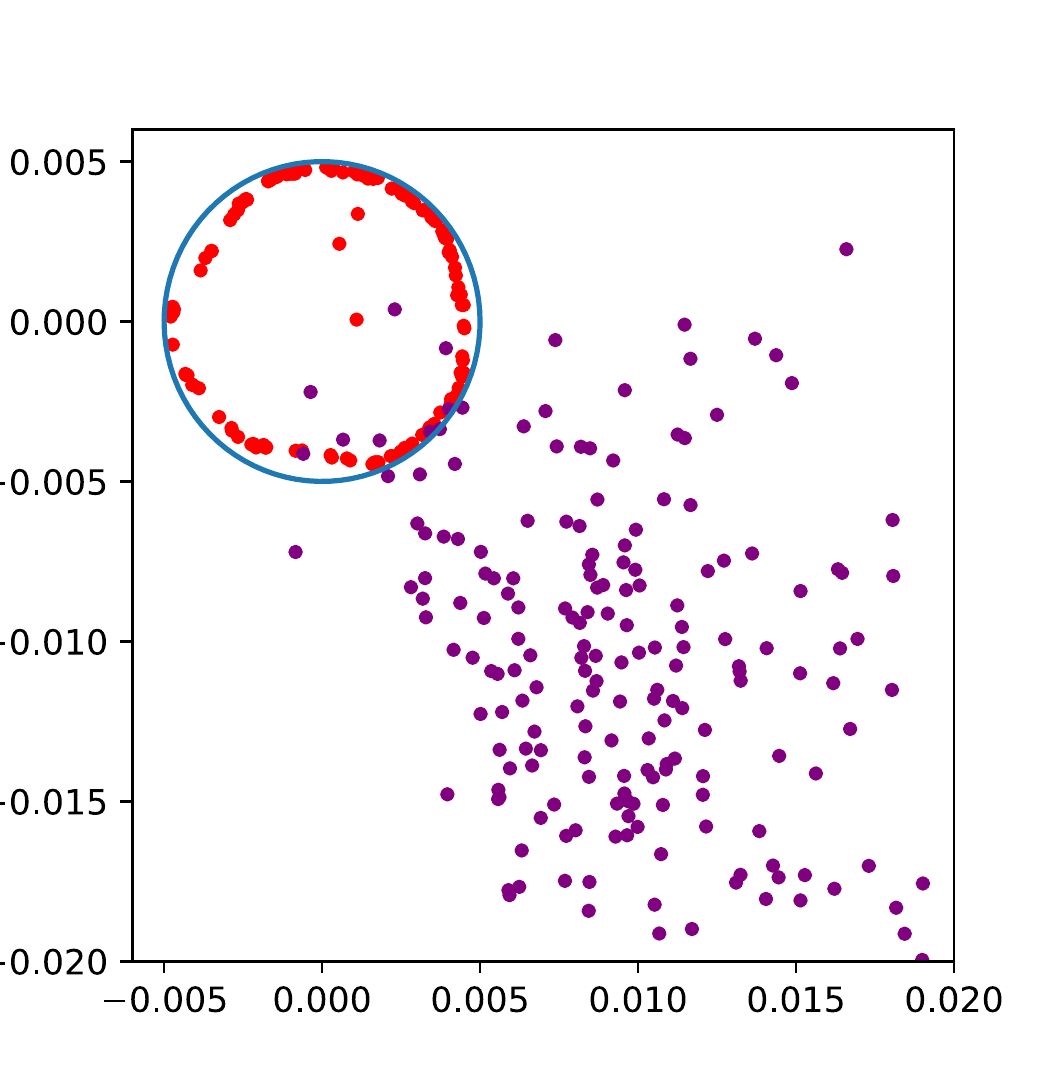}
    \caption{We plot the final positions of the inserted objects. Tests in our simulator (red) are confined within a small feasible region (a circle), meaning that no intersection has occurred, while most tests in Bullet (purple) are outside the circle.}
    \label{fig:datapoint}
\end{figure}

We then plot the $x$, $y$ coordinates of the obtained positions to show the horizontal locations of the inserted objects (See Fig. \ref{fig:datapoint}).
We draw a blue circle centered at $(0, 0)$ to prescribe the maximum range of positions such that the object will not intersect with the container. The radius of the circle equals the insertion margin $\Delta r = 0.005$. As shown in the figure, our method (red) leads to non-intersecting results for all the test cases. Furthermore, the positions are placed uniformly near the circle since both the object and the container are symmetric. On the contrary, the policy learned using Bullet (purple) contains a large offset from the origin, with $95\%$ of the points outside the circle, which indicates the occurrence of intersections in the test cases.

\subsection{Increasing Difficulty}
We further increase the difficulty of the peg-in-hole task by using a larger object ($r_1 = 0.039$), while other settings remain the same. Now the collision margin $\Delta r = 0.001$ is only $20\%$ in size compared to the previous one. We again obtain the control policy after 20 training iterations and execute it on 100 test cases. We observe that, for this more difficult task, our method achieves an $85\%$ success rate. This is still relatively high compared to the $91\%$ success rate from the previous task. On the other hand, PyBullet can no longer obtain meaningful results within the same number of training iterations after the increase in difficulty -- only 2 out of 100 test cases are successful.

\subsection{Summary}
In this section, we constructed a proof-of-concept RL environment to demonstrate that our simulator can be applied to train control policies for peg-in-hole tasks. The accurate contact feedback from Midas enables the agent to achieve better performance in fewer training iterations compared to Bullet. Midas is also able to handle tasks with extremely high precision since the contact threshold can be set to as small as the task needs for a common-size robot gripper and no interpenetration will happen in the simulation.

\section{Conclusion and Future Work}
We proposed Midas, a new robotics simulator based on previous works on the Incremental Potential Contact (IPC) model. Our method guarantees non-interpenetration for simulating multi-joint robots. We demonstrate the strengths of our method on various robot manipulation tasks with complicated contacts. We also test our simulator with a proof-of-concept reinforcement learning framework for peg-in-hole tasks. We confirm that our simulator is able to generate accurate control policies without artifacts.

As a potential future work, our simulator can be easily extended to simulate FEM soft bodies (e.g. rubber) and codimensional bodies (e.g. cloth and rod) as IPC supports automatic rigid-deformable coupling and arbitrarily thin geometries. We are determined to increase the capability of Midas and use it in more challenging robot manipulation tasks.

\addtolength{\textheight}{-12cm}   






\bibliographystyle{IEEEtran}
\bibliography{IEEEabrv,main}

\end{document}